\RequirePackage{amsmath}
\documentclass[runningheads]{llncs}
\usepackage[T1]{fontenc}
\usepackage{graphicx}
\usepackage{amsfonts}

\usepackage[square,numbers]{natbib}
\usepackage{xcolor}

\usepackage{float}

\usepackage{textgreek}

\usepackage{subcaption}
\usepackage{siunitx,tabularx,ragged2e,booktabs}
\newcolumntype{C}{>{\Centering}X}
\newcolumntype{T}[1]{S[table-format=#1]}

\usepackage{amsthm}

\usepackage{changepage}

\usepackage[subtle]{savetrees}

\begin{document}
\title{Achieving Group Fairness through Independence in Predictive Process Monitoring}
\titlerunning{Achieving Group Fairness through Independence in PPM}
\author{Jari Peeperkorn\orcidID{0000-0003-4644-4881} \and
Simon De Vos\orcidID{0000-0002-3032-8678}}
\authorrunning{Peeperkorn and De Vos}
\institute{Research Center for Information Systems Engineering (LIRIS), KU Leuven, Belgium \\
\email{jari.peeperkorn@kuleuven.be}}%
\maketitle              % typeset the header of the contribution
\begin{abstract}
Predictive process monitoring focuses on forecasting future states of ongoing process executions, such as predicting the outcome of a particular case. In recent years, the application of machine learning models in this domain has garnered significant scientific attention. When using historical execution data, which may contain biases or exhibit unfair behavior, these biases may be encoded into the trained models. Consequently, when such models are deployed to make decisions or guide interventions for new cases, they risk perpetuating this unwanted behavior. This work addresses group fairness in predictive process monitoring by investigating independence, i.e. ensuring predictions are unaffected by sensitive group membership. We explore independence through metrics for demographic parity such as $\Delta\mathit{DP}$, as well as recently introduced, threshold-independent distribution-based alternatives. Additionally, we propose a composite loss function existing of binary cross-entropy and a distribution-based loss (Wasserstein) to train models that balance predictive performance and fairness, and allow for customizable trade-offs. The effectiveness of both the fairness metrics and the composite loss functions is validated through a controlled experimental setup.

\end{abstract}
%
%
%

%The fairness here is group fairness. I also think a ``legal'' argument will probably fly best. What it is not: creating a fair process. The idea is that a company uses OOPPM to act (in earlier stages). This OOPPM model is trained on biased data. We don't want our predictor to reflect that, for legal/ethical reasons.

%Why not just take the top 10\% of each of the two binary features? You often have continuous data (not in batches), but more importantly here: we often need an early indicator for all of the running cases. 

%However, enforcing fairness often involves a trade-off with model performance, as mandating fairness across the entire output domain can inadvertently compromise a model’s predictive accuracy~\cite{kleinberg2016inherent, corbett2017algorithmic}. To cope with this, the results of our experiments express this fairness-performance trade-off as Pareto frontiers.  This approach provides a principled way to balance fairness against predictive utility, leaving the level of fairness enforcement as a policy choice.  Such decisions align with a risk-based approach where, depending on the application, different fairness tiers can be adhered to \simon{insert ref to EU AI Act?}.

\section{Introduction}

Predictive Process Monitoring (PPM) is a branch of process mining that aims to predict the future state of ongoing business processes based on historical event data. A specialized subset of PPM, known as Outcome-Oriented Predictive Process Monitoring (OOPPM), focuses on predicting specific outcomes or labels of process instances. By leveraging historical cases, OOPPM enables organizations to anticipate critical outcomes and intervene at earlier stages of the process, enabling earlier interventions that improve efficiency or reduce risks. Recent advancements in OOPPM have predominantly utilized machine and deep learning models trained on labeled historical data to achieve accurate predictions.

A significant ethical and legal challenge arises when these models are trained on biased data, which may encode systemic inequalities, such as those based on gender or ethnicity. Predictive models can inadvertently reproduce or even exacerbate these disparities if fairness considerations are not addressed. In recent years, different works have focused on using process mining to detect or discover such fairness concerns within a process~\cite{Pohl2023_workshoppaper, Muskan2025}. This work focuses on ensuring OOPPM models produce fair and unbiased predictions. While improving the fairness of the process execution itself is not the primary goal of these approaches, it is assumed that efforts are made in parallel to enhance the fairness of the process. Consequently, our approaches can then be used to measure and improve the extent to which the classifier's outcomes reflect these fairer executions, rather than perpetuating historical biases. For example, in a hiring process, if a model is trained on historically biased data, such as gender bias, it may carry that bias into its predictions, potentially reinforcing discrimination during early interventions. These biases can be direct, if the model decisions are (partially) based on protected attributes, or indirect, if decisions affect protected groups even without explicitly using their related attributes. From a legal and ethical perspective, tackling these is essential for organizations to align with fairness mandates, such as those outlined in regulations like the EU AI Act~\cite{european_ai_act_2021}.

A key fairness criterion discussed in this work is group fairness through independence, which ensures equal predictive outcomes across protected groups (e.g., defined by gender or ethnicity). Achieving fairness often involves trade-offs, as enforcing fairness constraints may reduce predictive performance~\cite{kleinberg2016inherent, corbett2017algorithmic}. To navigate this tension, this work evaluates the fairness-performance trade-offs, providing a principled framework for balancing these competing objectives. This approach allows stakeholders to make informed policy decisions about the desired level of fairness enforcement, aligning with a risk-based perspective that can vary depending on the application and its regulatory environment.
By addressing these challenges, this paper aims to contribute to the development of fair and effective OOPPM models that align predictive capabilities with ethical and legal fairness standards and to offer tools for practitioners and researchers to do so as well. To this extent, the main contributions of this work are:
\begin{itemize}
    \item Introducing group fairness into predictive process monitoring, with independence as the fairness criterion.
    \item Proposing and evaluating metrics for \emph{demographic parity} such as $\Delta$DP, alongside more advanced, threshold-independent alternatives \emph{area between probability density function curves} (ABPC) and \emph{area between cumulative density function curves} (ABCC).
    \item Incorporating \emph{integral probability metrics} (IPMs) into a composite loss function, complementing traditional loss functions such as binary cross-entropy. Experiments demonstrate that balancing IPMs with traditional loss functions enables flexible trade-offs between fairness and predictive accuracy.
\end{itemize}
The rest of this work is structured as follows: related work and notation are introduced in Section~\ref{sec:background}, followed by an in-depth introduction to the independence measures used in Section~\ref{sec:fairness}. The predictive setup is discussed in Section~\ref{sec:methodology}, followed by a proof-of-concept in~\ref{sec:experiments}. Some practical considerations and guidelines are considered in Section~\ref{sec:practical}, before concluding in section~\ref{sec:conclusion}. The adjustable and modular code is made available online, together with full experimental results\footnote{\url{https://github.com/jaripeeperkorn/Group-Fairness-in-Predictive-Process-Monitoring}}. 

\section{Background}
\label{sec:background}

\subsection{Preliminaries}
Executed \emph{activities} in a process are recorded as events in an event log \( L \). Each \emph{event} belongs to one \emph{case}, identified by its \emph{CaseID} \( c \in C \). An event \( e \) can be expressed as a tuple \( e = (c, a, t, d, s) \), where \( a \in A \) represents the \emph{activity} (i.e., the \emph{control-flow} attribute) and \( t \) is the timestamp of the event. Optionally, an event might have associated \emph{event-related attributes} \( \textbf{d} = (d_{1}, d_{2}, \dots, d_{m_{d}}) \), which are dynamic attributes that are event-specific (such as the resource executing the activity). Conversely, \emph{static attributes} \( \textbf{s} = (s_{1}, s_{2}, \dots, s_{m_{s}}) \) are case-level attributes that do not change during the execution of a case (such as customer information). A sequence of events that belong to a single case is referred to as a \emph{trace}. The outcome \( y \) of a trace is an attribute defined by the process owner, often binary in nature, indicating whether a specific criterion has been met~\cite{teinemaa2019outcome}. A \emph{prefix} is a portion of a trace, consisting of the first \( l \) events, where \( l \) is an integer smaller than the trace length. 
%In this work, we denote the \emph{protected groups} by the value of some (binary) sensitive attribute (a static case feature). 
In summary, we are working with a data set $\mathcal{D}=\left\{\left(\mathbf{x}_i, y_i, s_i\right)\right\}_{i=1}^N$. Here, $\mathbf{x}_i \in \mathbb{R}^d$ is the (encoded) input feature set for one sample, in this setup the full prefix sequence and $y_i$ is the true outcome of the case from which that prefix is derived. Let $s_i \in\{0,1\}$ be the sensitive attribute (i.e., a static case feature) value of sample $i$ that defines $\mathcal{S}_0=\left\{i: s_i=0\right\}$ and $\mathcal{S}_1=\left\{i: s_i=1\right\}$ as index sets of entities that, according to this sensitive feature $S$, belong to group $0$ or group $1$ respectively. %Both $\mathcal{S}_0$ and $\mathcal{S}_1$ could be protected groups (usually), or in specific cases one of the two. In this work, for consistency, we call both groups \textit{protected}, as this is closer to the goal of equal outcomes across groups.
We train a model $m: \mathbb{R}^d \rightarrow[0,1]$ that provides a propensity $\hat{y} \in [0,1]$. The model's output propensity scores may not always represent true probabilities, particularly if the model is not well-calibrated. In this work, however, due to its model-agnostic approach, we work with the direct output propensities. 

\subsection{Related Works}

Predictive process monitoring (PPM) addresses various tasks such as predicting the remaining time of a process~\cite{verenich2019survey}, identifying the next activity~\cite{tax2017predictive, Camargo2019}, or determining the final outcome of a process~\cite{teinemaa2019outcome, kratsch2021machine, Fransescomarino2019}. Different approaches in the literature range from using finite state machines~\cite{VanDerAalst2011} and stochastic Petri nets~\cite{Rogge-Solti2013} to machine learning techniques such as regression trees~\cite{deleoni2016} and ensembles~\cite{Spoel2013}. In recent years, lots of focus has been placed on deep learning techniques. Due to the natural fit with the sequential process data, recurrent neural networks~\cite{tax2017predictive, Camargo2019} and, more recently, transformer nets~\cite{Wuyts2024} have garnered a lot of attention. In addition to technical advancements in PPM, the importance of addressing fairness within process mining has gained attention. Fairness in process mining focuses on among others identifying and mitigating biases that may lead to discriminatory outcomes in processes or their analyses, within the broader context of AI regulatory compliance~\cite{Perry2022}. \cite{Pohl2023_workshoppaper} categorizes fairness concepts from machine learning, applies them to process mining, and highlights key fairness-related challenges for process mining. Other work proposes a fair classifier for root cause analysis in processes~\cite{Qafari_2019}, or an adapted genetic process discovery algorithm optimized for group fairness~\cite{Muskan2025}. Recently, \cite{pohl2023_paper} introduced a collection of simulated event logs designed to address the scarcity of fairness-aware datasets in process mining. Within PPM, integrating adversarial debiasing has been proposed to mitigate the influence of certain variables on biased predictions~\cite{deleoni_2024}, and different metrics have been explored as well~\cite{DeSilva2025}.

\section{Group Fairness through Independence}
\label{sec:fairness}

Predictive models used in PPM can exhibit discriminatory behavior toward specific groups. Such biases often stem from historical patterns of systematic disadvantage faced by certain populations. This work focuses on group-level fairness measures, like independence, which aim to equalize outcomes across protected groups, addressing systemic inequalities at the population level. Next to this, individual fairness, ensuring similar treatment for similar cases, provides a more granular perspective and requires different approaches such as generating counterfactuals~\cite{dwork2012fairness}. Features defining the protected group(s) may be present explicitly or implicitly. When explicitly included, removing sensitive attributes alone is still often insufficient to ensure fairness, as the principle of \textit{"fairness through unawareness"} has been shown to be ineffective~\cite{dwork2012fairness, Chen2019, barocas2023fairness}. Other features in the dataset could correlate with the \textit{sensitive} attribute, allowing its value to be inferred through these so-called proxy features, which can perpetuate bias. In PPM, such inference might stem from complex control-flow patterns in historical cases. Sometimes fully excluding these features can also lead to unexpected results, such as in the case of Simpsons's paradox (e.g., when gender correlates with physical attributes such as length)~\cite{Pearl2022}. Fairness in machine learning can be defined in various ways. Two other prominent approaches, other than independence, are separation and sufficiency. Separation corresponds to the idea of \emph{error rate parity} (for both false positive and false negative rates). Sufficiency, in short, means that predictions should be independent of the sensitive attribute given the actual outcome, ensuring \emph{calibration fairness}. Comprehensive reviews of these fairness definitions can be found in works such as~\cite{makhlouf2021applicability} and~\cite{barocas2023fairness}. For the rest of this work, we will focus on independence.

\subsection{Demographic Parity Metrics}
\label{sec:fairnessmetrics}

In machine learning, demographic or statistical parity is defined as independence on group-level~\citep{makhlouf2021applicability}.
This requires that predicted probabilities are independent of sensitive attributes. Commonly used metrics to measure demographic parity violations include the average propensity difference between the two groups, denoted as $\Delta\mathit{DP}_c$ (continuous) and defined by Eq.~\ref{eq:deltadp_c}, and the difference in the proportion of positive predictions between the two groups, dubbed $\Delta\mathit{DP}_b^t$ (binary at threshold $t$) and defined by Eq.~\ref{eq:deltadp_b} \cite{Zemel2013, Zeng2022, Kim2022, han2023}. Here, $\mathbf{1}(\hat{y}_n > t)$ is the indicator function that equals $1$ if the predicted value exceeds threshold $t$ and $0$ otherwise.

\begin{equation}\label{eq:deltadp_c}
\Delta D P_c=\left|\frac{\sum_{n \in \mathcal{S}_0} \hat{y}_n}{|\mathcal{S}_0|}-\frac{\sum_{n \in \mathcal{S}_1} \hat{y}_n}{|\mathcal{S}_1|}\right|
\end{equation}

\begin{equation}\label{eq:deltadp_b}
\Delta D P_b^t=\left|\frac{\sum_{n \in \mathcal{S}_0} \mathbf{1}(\hat{y}_n > t)}{|\mathcal{S}_0|}-\frac{\sum_{n \in \mathcal{S}_1} \mathbf{1}(\hat{y}_n > t)}{|\mathcal{S}_1|}\right|
\end{equation}

To clarify consider an early intervention PPM model in hiring, where a process is automatically accepted or rejected if the prediction exceeds or falls below a threshold $t$. $\Delta D P_b^t$ measures the difference in intervention rates between e.g. gender groups, while $\Delta D P_c$ captures the difference in the average model propensity for both groups. A limitation of $\Delta\mathit{DP}_b^t$ is its dependence on the classification threshold $t$. Threshold-sensitive fairness metrics might not fully capture disparities across the entire range of outputs, which is particularly important in dynamic settings like PPM where $t$ can vary due to application-specific properties such as changing cost-benefit ratios or scarce resources.
Such settings require distribution-based approaches to ensure flexibility post-training or even post-deployment. For example, when optimizing profit, cost-sensitive thresholding has been demonstrated to be effective~\cite{vanderschueren2022predict}.
To address these limitations, recent work has shifted towards evaluating fairness across the entire output distribution. This approach allows predictive models independently of the chosen threshold~\cite{han2023}. Therefore, following~\cite{han2023}, we adopt the metrics \emph{area between probability density function curves} (ABPC) and \emph{area between cumulative density function curves} (ABCC). \\

\noindent\begin{minipage}{.5\linewidth}
\begin{equation}\label{eq:abpc}
\operatorname{ABPC}=\int_0^1\left|f_0(x)-f_1(x)\right| \mathrm{d} x
\end{equation}
\end{minipage}%
\begin{minipage}{.5\linewidth}
\begin{equation}\label{eq:abcc}
\operatorname{ABCC}=\int_0^1\left|F_0(x)-F_1(x)\right| \mathrm{d} x
\end{equation} 
\end{minipage} \\

Here, $f_0(\cdot)$ and $f_1(\cdot)$ represent the probability density functions (PDFs) of protected groups 0 and 1, respectively, and $F_0(\cdot)$ and $F_1(\cdot)$ denote their corresponding cumulative density functions (CDFs). ABPC values range between $0$ (full parity) and $2$, and ABCC values between $0$ and $1$. Since we are working with finite samples, we make estimations in practice. For ABPC we estimate the PDFs via kernel density estimation (KDE)~\cite{Rosenblatt1956}, and for ABCC directly use the empirical CDFs~\cite{wasserman2006}, following~\cite{han2023}. Both metrics are calculated using the composite trapezoidal rule\footnote{Sufficient precision is ensured by using 10,000 steps per integration~\cite{han2023}.}.

\begin{figure}
\centering
\begin{subfigure}[h]{0.49\textwidth}
\includegraphics[width=\textwidth]{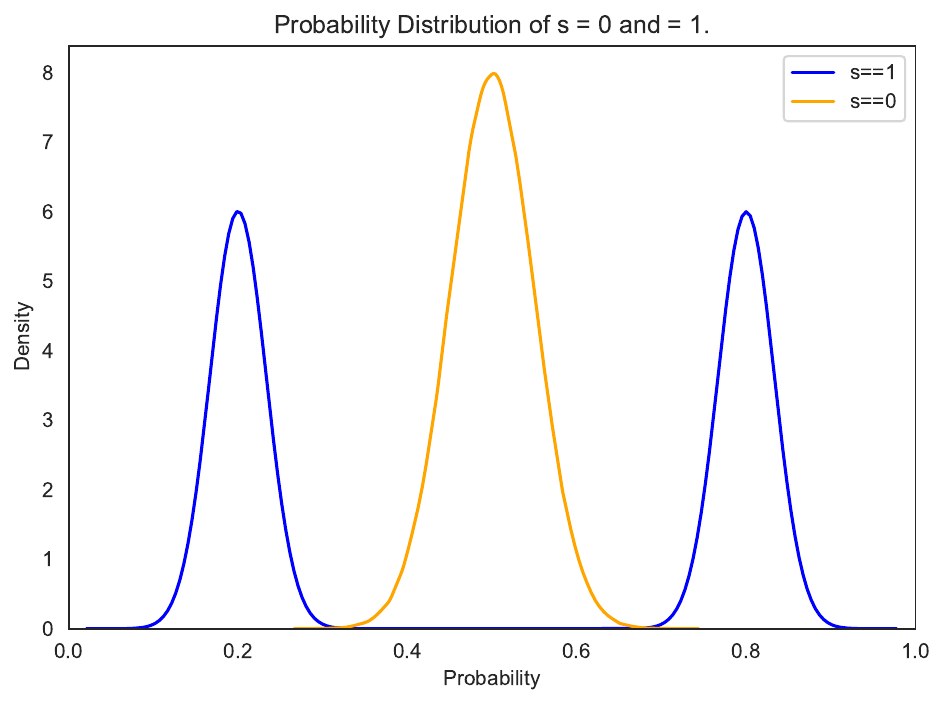}
\caption{Propensity distributions for $\mathcal{S}_0$ and $\mathcal{S}_1$}
\end{subfigure}
\hfill
\begin{subfigure}[h]{0.49\textwidth}
\includegraphics[width=\textwidth]{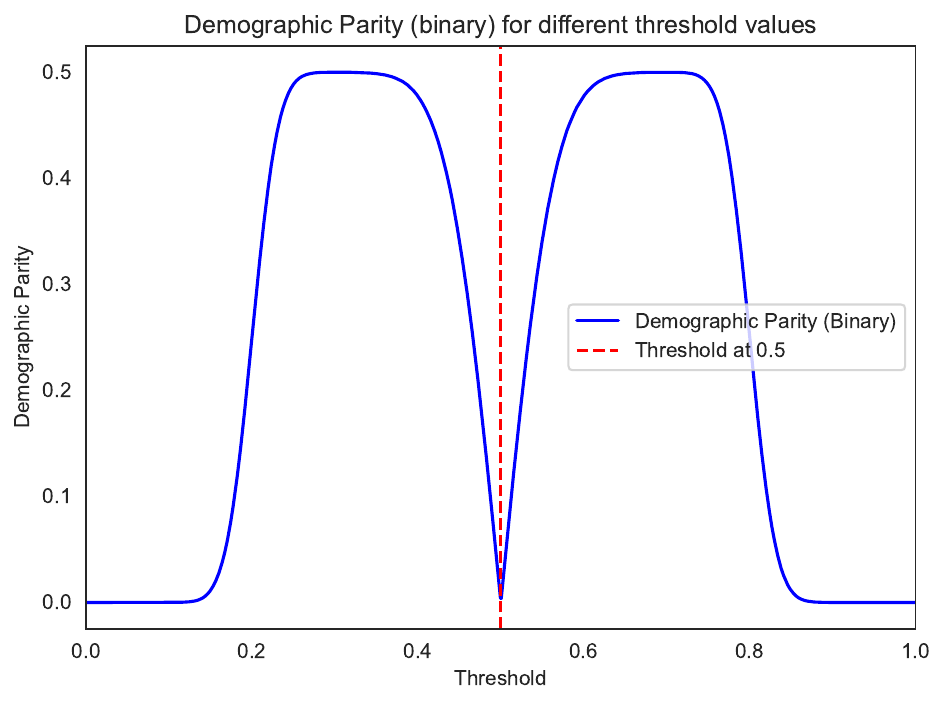}
\caption{$\Delta\mathit{DP}_b^t$ scores in function of $t$}
\end{subfigure}%
\caption{A toy example showing the need for threshold-free independence metrics.}
\label{fig:toy_example}
\end{figure}

Figure~\ref{fig:toy_example} demonstrates the advantage of using distribution-based metrics, next to their threshold-independence. The left subplot shows the propensity distributions for two groups (e.g. for male and female candidates in a hiring process). Although $\Delta\mathit{DP}_c = 0$, the distributions are clearly not independent of the sensitive attribute, as reflected by ABPC and ABCC values of 2.0 and 0.26, respectively. The right subplot illustrates how $\Delta\mathit{DP}_b^t$ varies significantly with threshold $t$, highlighting its sensitivity to the choice of $t$.

\subsection{Integral Probability Metric Loss}

To incorporate fairness as defined above into the learning process, we include IPMs~\cite{Muller1997, sriperumbudur2012empirical} to quantify the distance between the two prediction distributions $p(\hat{y}|s=0)$ and $p(\hat{y}|s=1)$. IPMs have been earlier applied in machine learning to, e.g., match output and true probability densities of binary classifiers~\cite{sriperumbudur2009}, or to learn balanced covariate representations in causal inference~\cite{Shalit2017}. Additionally, they have been recently employed to improve fairness metrics like demographic parity~\cite{Kim2022}. During training, IPMs are incorporated into the model's objective function through a composite loss formulation:

\begin{equation}
\label{eq:IPM}
    \mathcal{L}_{\text{total}} = (1-\lambda) \cdot \mathcal{L}_{\text{BCE}} + \lambda \cdot \mathcal{L}_{\text{IPM}}
\end{equation}

Here, \( \mathcal{L}_{\text{BCE}} \) denotes a standard supervised learning loss, such as binary cross-entropy (BCE), which measures the predictive performance of the model. On the other hand, \( \mathcal{L}_{\text{IPM}} \) penalizes discrepancies between the prediction distributions across protected groups. The hyperparameter \( \lambda \in [0, 1] \) governs the trade-off between maximizing predictive accuracy and ensuring fairness. By incorporating \( \mathcal{L}_{\text{IPM}} \) into the training process, the model is guided not only to minimize traditional predictive losses but also to satisfy a soft fairness constraint, encouraging more equal outcomes across protected groups. In the hiring process example, \( \mathcal{L}_{\text{BCE}} \) trains early predictions to match (potentially biased) outcomes, while \( \mathcal{L}_{\text{IPM}} \) encourages similar propensity distributions across groups. Adding a penalty proportional to the difference between the two propensity distributions nudges the model toward aligning the distributions more closely, thereby reducing independence violations. In this work, we employ the Wasserstein distance (also known as Earth Mover's Distance) as the IPM. The Wasserstein distance is a measure of the effort required to transform one probability distribution into another, taking into account where (at which probabilities) in the distribution the differences occur, i.e., if further apart, the effort should be higher. The optimal mass transformation from one distribution to the other is determined by the area between the cumulative distribution functions. In this way, using Wasserstein as IPM loss is more aligned with optimizing ABCC than with optimizing ABPC. For a formal definition, the reader is referred to~\cite{villani2009optimal}. The Wasserstein distance provides a geometrically intuitive and robust way to measure the distributional discrepancy, making it well-suited for enforcing group fairness. In this work, the more efficient Sinkhorn approximation is used, as implemented by~\cite{Shalit2017}, allowing it to be used inside the training loop as well.
Alternatives to measure distributional differences include Kullback-Leibler (KL) Divergence~\cite{boyd2004convex} and the kernel-based Maximum Mean Discrepancy (MMD)~\cite{gretton2012}.

\section{Methodology}
\label{sec:methodology}

\subsection{Preprocessing}
The event logs are divided into training and test sets at the case level, with an 80-20\% split. While fully correct practice involves removing overlapping cases and biased cases near the end of the test set (as described in~\cite{Weytjens2022}), this step is omitted due to the use of artificially simulated data and the focus on conceptual demonstrations rather than predictive performance. Cases are converted into prefix-outcome pairs. Prefixes are handled by defining a maximum length: shorter prefixes are right-padded, and longer ones are left-truncated. Outcomes are defined based on the presence of specific activities in the process executions. Only prefixes up to but not including the target activity are included in the dataset. A sensitive feature is selected to determine the protected groups. The implementation allows for including or excluding this feature from the input space. Static case features are included as input at every event in a case. Numerical features are min-max scaled to the range $[0,1]$, while non-binary categorical features (e.g., activity, resource) are one-hot encoded and passed through separate embedding layers. A validation set is used for hyperparameter tuning, early stopping, and threshold tuning. This validation set is created by splitting 20\% of the prefix-outcome pairs from the training samples.

\subsection{Model}

\begin{figure}
    \centering
    \includegraphics[width=0.90\textwidth]{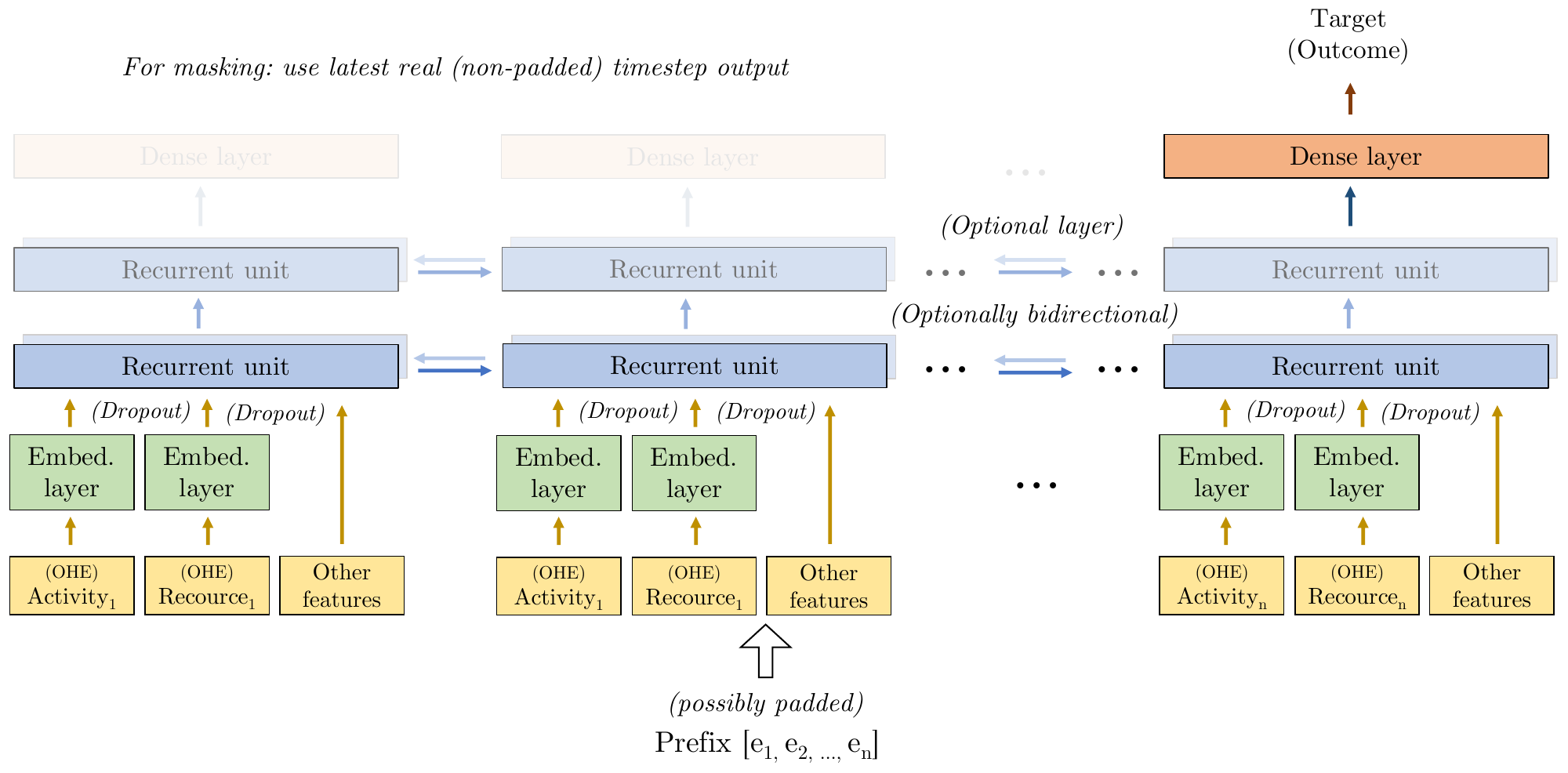}
    \caption{A graphical depiction of the LSTM model used.}
    \label{fig:LSTM_model}
\end{figure}

An LSTM model~\cite{LSTM} was selected for these conceptual experiments, due to its widespread use in PPM research and its natural suitability for sequential data. However, the metrics and adjusted loss functions proposed in this work are compatible with various other models. A graphical representation of the LSTM classifier used in our experiments is shown in Figure~\ref{fig:LSTM_model}. Categorical features are passed through individual trainable embedding layers with adjustable sizes. At each timestep, the outputs of these embeddings are combined with binary and numerical features to form the input to a recurrent LSTM layer, with dropout applied for regularization. The LSTM layer can be configured as bidirectional (processing the sequence in both forward and backward directions), and multiple layers can be stacked. The output from the final LSTM unit is fed into a dense layer to produce a binary propensity. To handle right-padded sequences, masking is employed, ensuring that the output of the LSTM corresponds to the last valid (unpadded) event.

The models are trained using the AdamW optimizer~\cite{Loshchilov2017}. A learning rate scheduler is applied to reduce the learning rate by a factor of $0.75$ every 10 epochs without improvement (greater than 0.001). Early stopping based on validation loss is employed, with a patience setting of 20 for hyperparameter tuning and 50 for the main experiments. Training is capped at 300 epochs. The loss function used is BCE, or, if $\lambda > 0$, a combination of BCE and the Wasserstein distance as IPM (as defined in eq.~\ref{eq:IPM}).

\paragraph{Hyperparameters}
The model has several hyperparameters that were optimized via grid search, including: the number of LSTM layers (1 or 2), the bidirectionality of this layer, and its size (16, 32 or 64), the mini-batch size (128, 256, 512), the starting learning rate (0.0001 or 0.001) and the dropout (0.2 or 0.4). The embedding layer size is set to the square root of the vocabulary size for each categorical feature. The grid search is performed once per event log using the BCE loss, and hyperparameters are selected using the validation set Area-under-the-curve (AUC) score. These are subsequently used in all experiments involving the corresponding event log. For experiments involving the IPM-composite loss functions, it was opted to keep the batch size fixed at 512 to ensure reliable computation of batch-level statistics.

\section{Experimental Setup \& Results}
\label{sec:experiments}

To validate the proposed metrics and the use of the composite loss, we conducted proof-of-concept experiments. These are not aimed to showcase predictive performance, but to demonstrate the potential of the fairness metrics and IPM loss, which are modular and adaptable across models. Experiment 1 demonstrates how the various independence metrics described above can be applied in a predictive process monitoring context, highlighting their potential differences. Experiment 2 illustrates how independence can be improved by incorporating the IPM loss.

\subsection{Datasets}
To ensure a controlled setup, we utilized several artificial datasets introduced in~\cite{pohl2023_paper}, which were designed to represent varying degrees of discrimination~\cite{Pohl2023}. These event logs were selected because of the different levels of bias they display within the same process. Since the focus of this experiment is to demonstrate the trade-off between independence and predictive performance, the selected event logs are well-suited despite their artificial nature. Their varying levels of bias within the same process effectively showcase how to measure and navigate this trade-off. Three processes were selected from the available event logs\footnote{The hospital process was excluded because its outcome variable did not correlate directly with the sensitive features.}. The first is a hiring process, which simulates recruitment workflows where the successful outcome is marked by the \textit{Make Job Offer} activity. The second process concerns loan applications, where a positive outcome corresponds to cases containing the \textit{Sign Loan Agreement} activity. The third process involves rental applications, with the \textit{Sign Contract} activity indicating a successful outcome. Some process executions include events occurring after the outcome-determining activity; these were excluded when creating prefixes in the dataset. \\

\begin{table*}[htbp!]
\centering
\caption{Information on the event logs~\cite{Pohl2023}, sensitive feature is \textit{protected}.}
\label{tab:overview_logs}
\begin{tabularx}{\textwidth}{l>{\centering\arraybackslash}X>{\centering\arraybackslash}X>{\centering\arraybackslash}X>{\centering\arraybackslash}X>{\centering\arraybackslash}X>
{\centering\arraybackslash}X>
{\centering\arraybackslash}X>{\centering\arraybackslash}X>{\centering\arraybackslash}X>{\centering\arraybackslash}X}
\toprule
 &
\multicolumn{5}{c}{Training (\& Validation) Set} &
\multicolumn{5}{c}{Test Set} \\
\cmidrule(lr){2-6} \cmidrule(lr){7-11}
 {Name} & 
 \#Pref. & \%+ & \% $\mathcal{S}_1$ & $\% \mathcal{S}_0^+$ & $\% \mathcal{S}_1^+$ &
 \#Pref. & \%+ & \% $\mathcal{S}_1$ & $\% \mathcal{S}_0^+$ & $\% \mathcal{S}_1^+$ \\
 \midrule
hiring\_high & 40268 & 41.24 & 19.94 & 48.86 & 10.66 & 10090 & 40.58 & 20.46 & 48.02 & 11.68 \\
hiring\_medium & 43753 & 46.05 & 15.71 & 50.61 & 21.58 & 10983 & 46.04 & 16.59 & 50.54 & 23.44 \\
hiring\_low & 45633 & 50.01 & 9.24 & 51.44 & 35.97 & 11566 & 50.07 & 9.43 & 51.27 & 38.50 \\ \hline
lending\_high & 36999 & 24.51 & 30.44 & 32.33 & 6.63 & 9196 & 25.13 & 32.11 & 33.69 & 7.04 \\
lending\_medium & 36582 & 27.64 & 21.16 & 31.51 & 13.23 & 9101 & 28.48 & 21.02 & 32.62 & 12.91 \\
lending\_low & 37994 & 30.52 & 9.78 & 31.73 & 19.35 & 9480 & 30.75 & 9.48 & 31.33 & 25.25 \\ \hline
renting\_high & 40516 & 28.91 & 25.67 & 34.49 & 12.78 & 10132 & 30.72 & 26.32 & 36.33 & 15.04 \\
renting\_medium & 42089 & 42.21 & 9.68 & 43.56 & 29.66 & 10533 & 41.37 & 10.29 & 43.17 & 25.65 \\
renting\_low & 41723 & 34.87 & 18.06 & 37.91 & 21.08 & 10413 & 34.50 & 18.79 & 37.54 & 21.41 \\
\bottomrule
\end{tabularx}
\end{table*}

Each event log comprises 10,000 cases. In addition to activity labels and resource event features, each log contains various binary features that could serve as sensitive attributes, such as \textit{gender} and \textit{religious affiliation}, as well as a \textit{protected} feature that identifies the protected groups. Some logs also include continuous features, such as \textit{age} or \textit{years of education}. For the experiments presented in this paper, the \textit{case:protected} feature was used as the sensitive attribute. However, experiments were also conducted using other binary features (static or case-level); the results of these additional experiments are available in the online repository. Table~\ref{tab:overview_logs} provides an overview of the event logs, including the number of prefixes in both the training and test datasets, the percentage of prefixes with positive (successful) outcomes, and the percentage of prefixes corresponding to cases belonging to the protected group $\mathcal{S}_1$, corresponding to cases for which \textit{case:protected} = \textit{True}. Additionally, the metrics $\%\mathcal{S}_0^+$ and $\% \mathcal{S}_1^+$ denote the percentages of positive outcomes within the two groups. For all of the experiments we kept the maximum prefix length at $6$.

\subsection{Experiment 1: Assessing Fairness Metrics}

The first experiment demonstrates the use of various independence fairness metrics within the OOPPM framework. An LSTM outcome classifier is trained for each event log from Table~\ref{tab:overview_logs}, and evaluated on the test set. The evaluation included standard performance metrics AUC, accuracy, F1 scores, as well as the fairness metrics introduced in Section~\ref{sec:fairnessmetrics}. For threshold-dependent metrics, results were computed for a fixed threshold of 0.5 ($\text{F1}_{0.5}$, $\text{Acc}_{0.5}$ and $\Delta\mathit{DP}_b^{0.5}$) and the threshold maximizing the F1 score on the validation set ($\text{F1}_{\text{opt.}}$, $\text{Acc}_{\text{opt.}}$ and $\Delta\mathit{DP}_b^{\text{opt.}}$). This experiment not only introduces independence metrics to OOPPM but also highlights the advantages of distribution-based metrics like ABPC and ABCC. Two settings were tested: one where the sensitive feature was included in the input space, and another where it was removed.

\begin{table*}[htpb!]
\centering
\caption{Results Exp. 1, sensitive feature is \textit{protected} (no removal).}
\label{tab:results_exp1}
\resizebox{0.99\linewidth}{!}{%
\begin{tabular}{l|ccccc|ccc|cc}
\toprule
Log & AUC &  $\text{F1}_{0.5}$ &  $\text{F1}_\text{Opt.}$ &$\text{Acc.}_{0.5}$ &  $\text{Acc.}_\text{Opt.}$ & $\Delta\mathit{DP}_b^{0.5}$ & $\Delta\mathit{DP}_b^{\text{opt.}}$ & $\Delta\mathit{DP}_\text{c}$ & ABPC & ABCC \\
\midrule
{hiring\_high} & 0.75 & 0.56 & 0.64 & 0.71 & 0.57 & 0.25 & 0.90 & 0.38 & 1.71 & 0.38 \\
{hiring\_medium} & 0.72 & 0.58 & 0.66 & 0.69 & 0.55 & 0.17 & 0.80 & 0.27 & 1.66 & 0.27 \\
{hiring\_low} & 0.70 & 0.59 & 0.67 & 0.67 & 0.53 & 0.05 & 0.69 & 0.14 & 1.30 & 0.14 \\ \hline
{lending\_high} & 0.71 & 0.00 & 0.53 & 0.75 & 0.60 & 0.00 & 0.89 & 0.24 & 1.85 & 0.24 \\
{lending\_medium} & 0.65 & 0.00 & 0.52 & 0.72 & 0.53 & 0.00 & 0.89 & 0.18 & 1.81 & 0.18 \\
{lending\_low} & 0.59 & 0.07 & 0.51 & 0.68 & 0.43 & 0.02 & 0.22 & 0.11 & 1.02 & 0.11 \\ \hline
{renting\_high} & 0.65 & 0.12 & 0.53 & 0.69 & 0.55 & 0.06 & 0.81 & 0.22 & 1.70 & 0.22 \\
{renting\_medium} & 0.61 & 0.28 & 0.61 & 0.59 & 0.52 & 0.17 & 0.42 & 0.13 & 1.51 & 0.13 \\
{renting\_low} & 0.63 & 0.00 & 0.55 & 0.65 & 0.53 & 0.00 & 0.86 & 0.13 & 1.66 & 0.13 \\
\bottomrule
\end{tabular}%
}
\end{table*}

\begin{figure}[htpb!]
\begin{adjustwidth}{-2.0cm}{-2.0cm}
\centering
\begin{subfigure}[htbp]{0.33\linewidth}
\includegraphics[width=\textwidth]{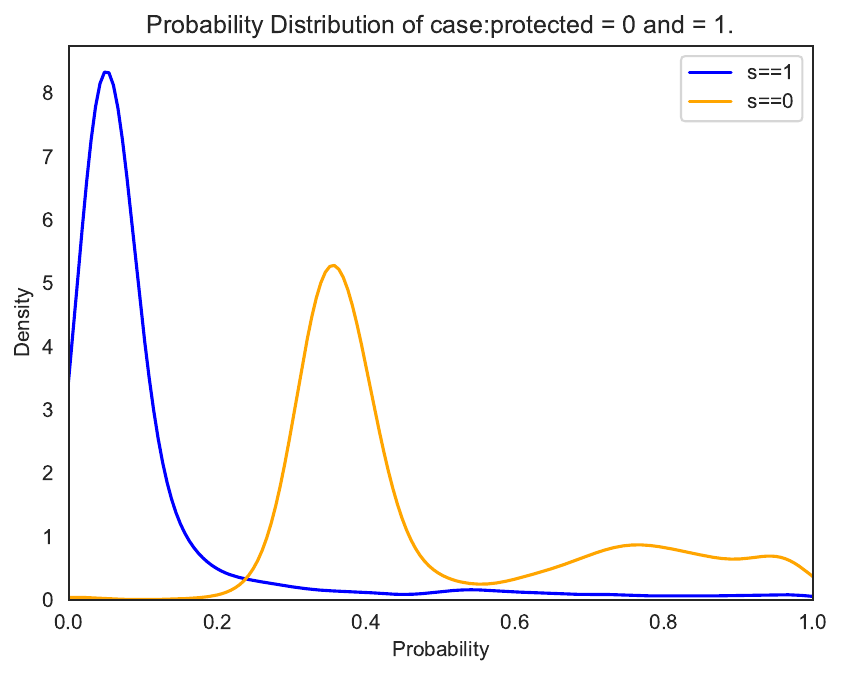}
\caption{hiring\_high}
\end{subfigure}
\begin{subfigure}[htbp]{0.33\linewidth}
\includegraphics[width=\textwidth]{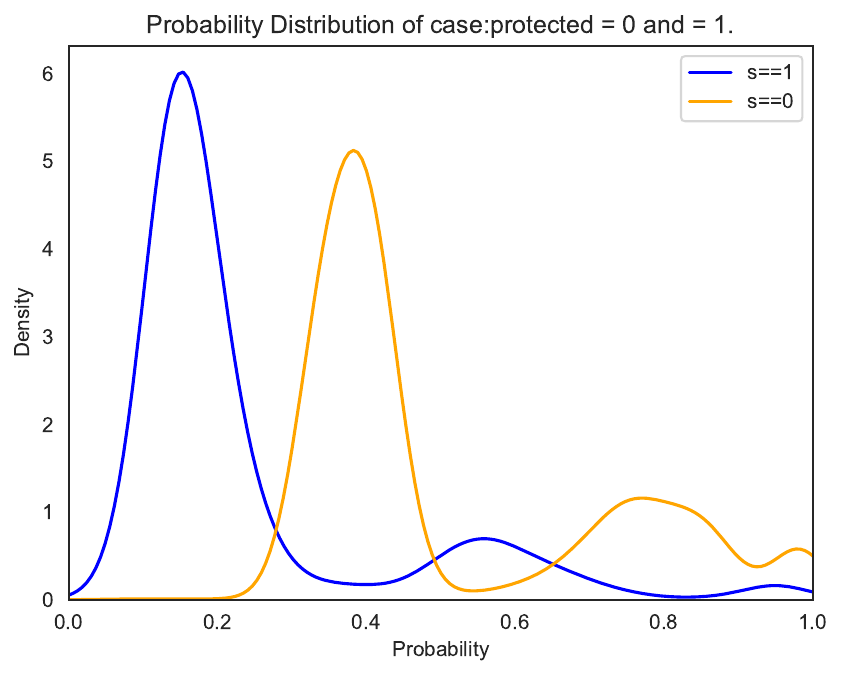}
\caption{hiring\_medium}
\end{subfigure}% 
\begin{subfigure}[htbp]{0.33\linewidth}
\includegraphics[width=\textwidth]{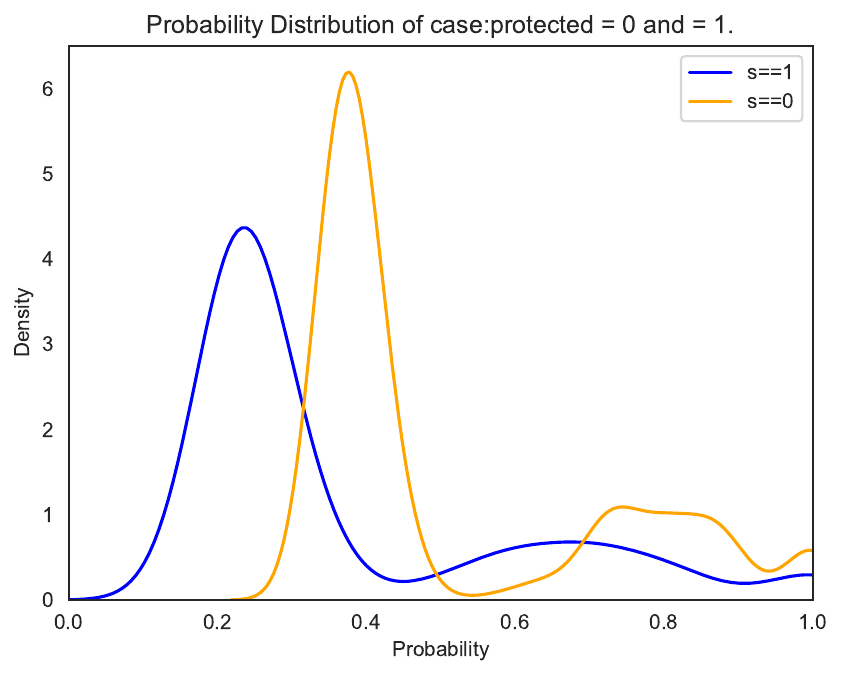}
\caption{hiring\_low}
\end{subfigure}
\end{adjustwidth}
\caption{The LSTM propensity densities for the $hiring$ event log.}
\label{fig:distr_plots}
\end{figure}

Table~\ref{tab:results_exp1} displays results for models trained with sensitive features included. While this may not always reflect realistic use cases, the goal here was to explore the metrics' ability to measure demographic parity violations, even under conditions where the models are explicitly trained on biased data. For the hiring process we can see a clear trend:  as bias in the event log decreases, both predictive scores and demographic parity violations reduce. For both the renting and lending logs these trends are a little less pronounced. The renting\_medium the DP metrics indicate less bias than renting\_low, but this is in line with the higher $\mathcal{S}_1^+$ value in Table~\ref{tab:overview_logs}. Some classifiers produced low accuracy at F1-tuned thresholds, while others resulted in F1 scores of zero at $t=0.5$, as all samples' propensity scores $\hat{y}$ fell either above or below the threshold, therefore also resulting in $\Delta\mathit{DP}_b^{0.5} = 0$. The scores of $\Delta\mathit{DP}_b^{\text{opt.}}$ are also noticeably larger than $\Delta\mathit{DP}_b^{0.5}$. Figure~\ref{fig:distr_plots} visualizes propensity distributions for hiring process classifiers across varying bias levels. The distributions show increased overlap, i.e., predictions are more fair, as bias in the event log, and consequently in the classifier, decreases.

\begin{table*}[htbp!]
\centering
\caption{Results Exp. 1, sensitive feature is \textit{protected} (with removal).}
\label{tab:results_exp1-removal}
\resizebox{0.99\linewidth}{!}{%
\begin{tabular}{l|ccccc|ccc|cc}
\toprule
Log & AUC &  $\text{F1}_{0.5}$ &  $\text{F1}_\text{Opt.}$ &$\text{Acc.}_{0.5}$ &  $\text{Acc.}_\text{Opt.}$ & $\Delta\mathit{DP}_b^{0.5}$ & $\Delta\mathit{DP}_b^{\text{opt.}}$ & $\Delta\mathit{DP}_\text{c}$ & ABPC & ABCC \\
\midrule
{hiring\_high} & 0.73 & 0.56 & 0.62 & 0.71 & 0.57 & 0.23 & 0.61 & 0.22 & 1.11 & 0.22 \\
{hiring\_medium} & 0.71 & 0.58 & 0.64 & 0.69 & 0.59 & 0.15 & 0.40 & 0.13 & 0.75 & 0.13 \\
{hiring\_low} & 0.7 & 0.59 & 0.67 & 0.67 & 0.52 & 0.07 & 0.46 & 0.10 & 0.87 & 0.10 \\ \hline
{lending\_high} & 0.66 & 0.00 & 0.46 & 0.75 & 0.54 & 0.00 & 0.49 & 0.09 & 0.99 & 0.09 \\
{lending\_medium} & 0.62 & 0.00 & 0.49 & 0.72 & 0.49 & 0.00 & 0.45 & 0.04 & 0.84 & 0.04 \\
{lending\_low} & 0.59 & 0.08 & 0.51 & 0.68 & 0.42 & 0.03 & 0.02 & 0.07 & 0.73 & 0.07 \\ \hline
{renting\_high} & 0.60 & 0.02 & 0.49 & 0.69 & 0.43 & 0.00 & 0.06 & 0.03 & 0.40 & 0.03 \\
{renting\_medium} & 0.61 & 0.29 & 0.61 & 0.59 & 0.51 & 0.14 & 0.25 & 0.07 & 0.93 & 0.07 \\
{renting\_low} & 0.61 & 0.00 & 0.54 & 0.65 & 0.46 & 0.00 & 0.07 & 0.03 & 0.39 & 0.03 \\
\bottomrule
\end{tabular}%
}
\end{table*}

For completeness, we also include results where the sensitive feature was excluded from the dataset, as shown in Table~\ref{tab:results_exp1-removal}. These results reveal significantly lower demographic parity violation scores, at the cost of slightly reduced predictive performance. However, since the protected cases in the event log generation were selected based on other features (e.g., gender), sensitive information persists indirectly through these proxy variables and DP metrics do not fall to $0$. This mirrors real-world scenarios where biases often remain embedded in correlated attributes, even after explicit sensitive features are removed. In this experimental setup, removing all binary features to eliminate indirect bias would leave the dataset too sparse for meaningful prediction. 

\subsection{Experiment 2: Testing IPM Loss}

\begin{figure}
\begin{adjustwidth}{-2cm}{-2cm}
\centering
\begin{subfigure}[htbp]{0.49\linewidth}
\includegraphics[width=\textwidth]{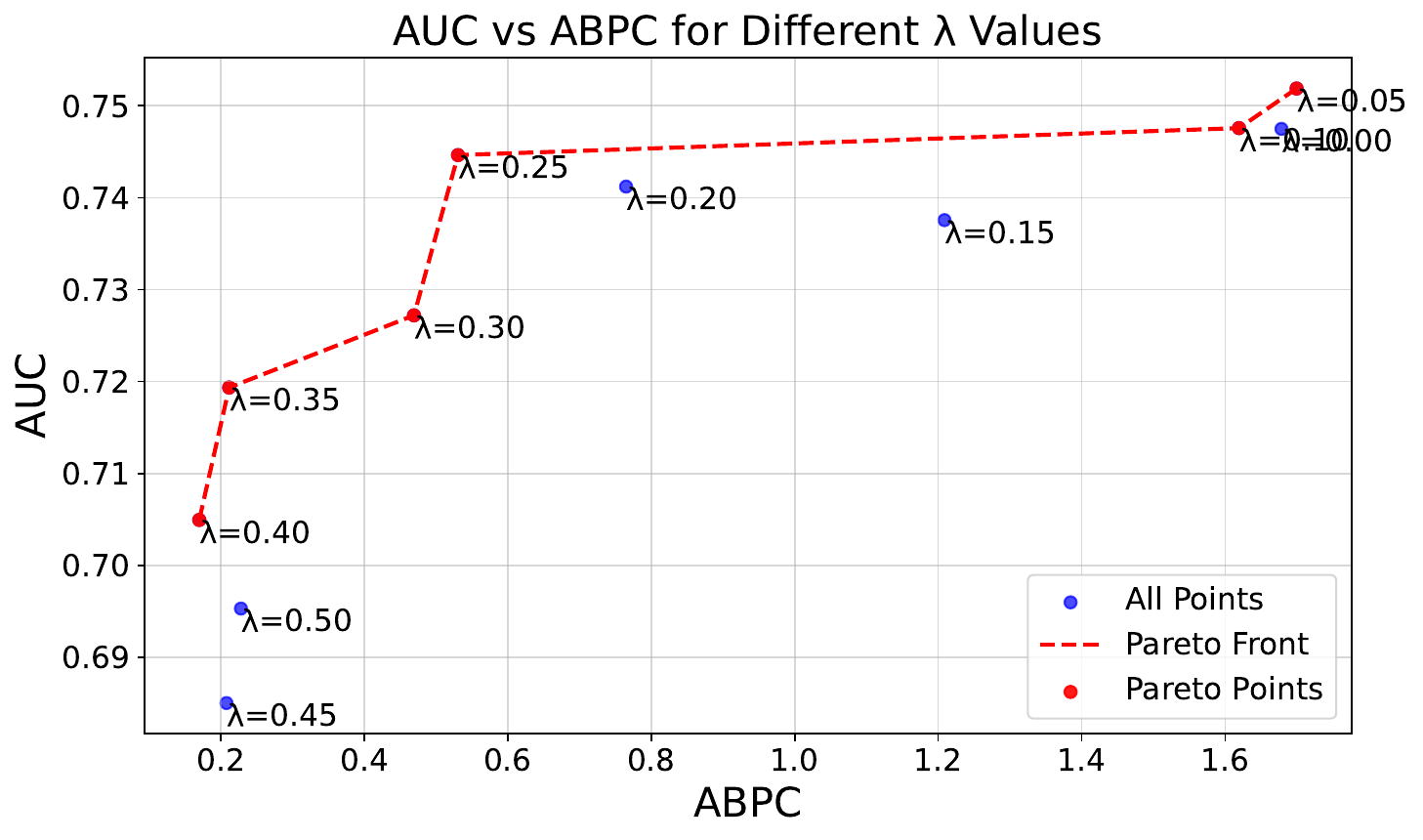}
\caption{AUC vs ABPC of hiring\_high log.}
\end{subfigure}
\begin{subfigure}[htbp]{0.49\linewidth}
\includegraphics[width=\textwidth]{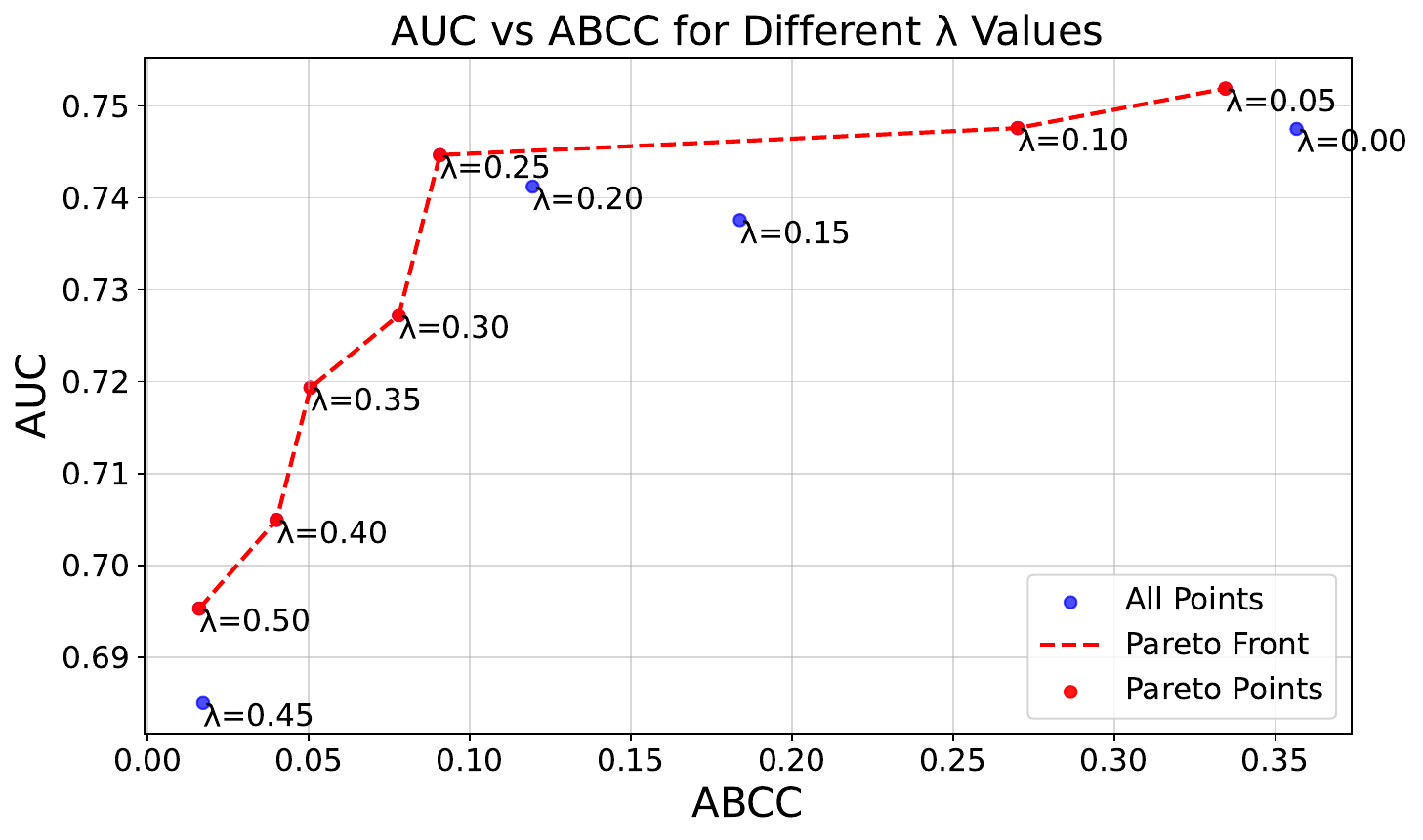}
\caption{AUC vs ABCC of hiring\_high log.}
\end{subfigure}% \\

\begin{subfigure}[htbp]{0.49\linewidth}
\includegraphics[width=\textwidth]{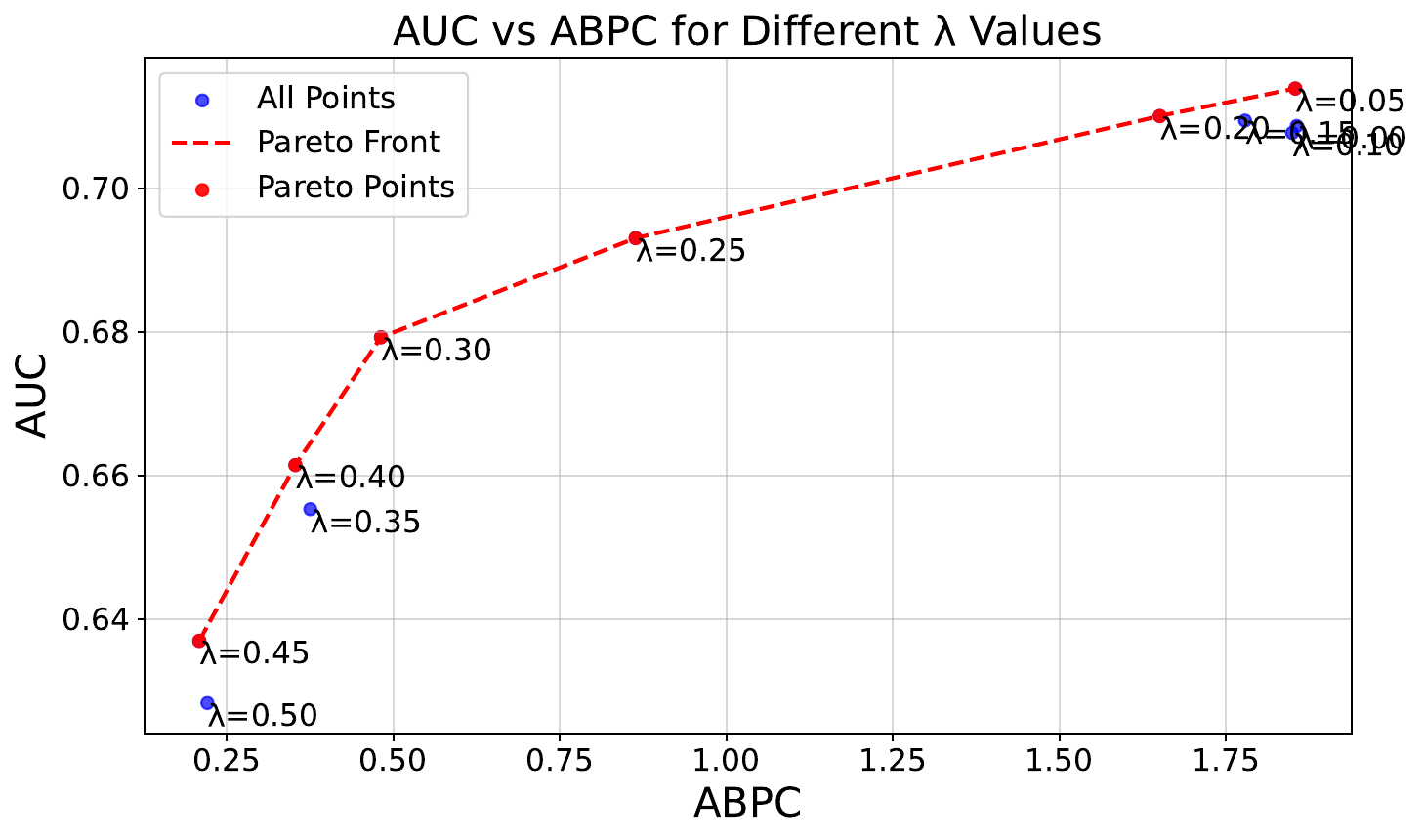}
\caption{AUC vs ABPC of lending\_high log.}
\end{subfigure}
\begin{subfigure}[htbp]{0.49\linewidth}
\includegraphics[width=\textwidth]{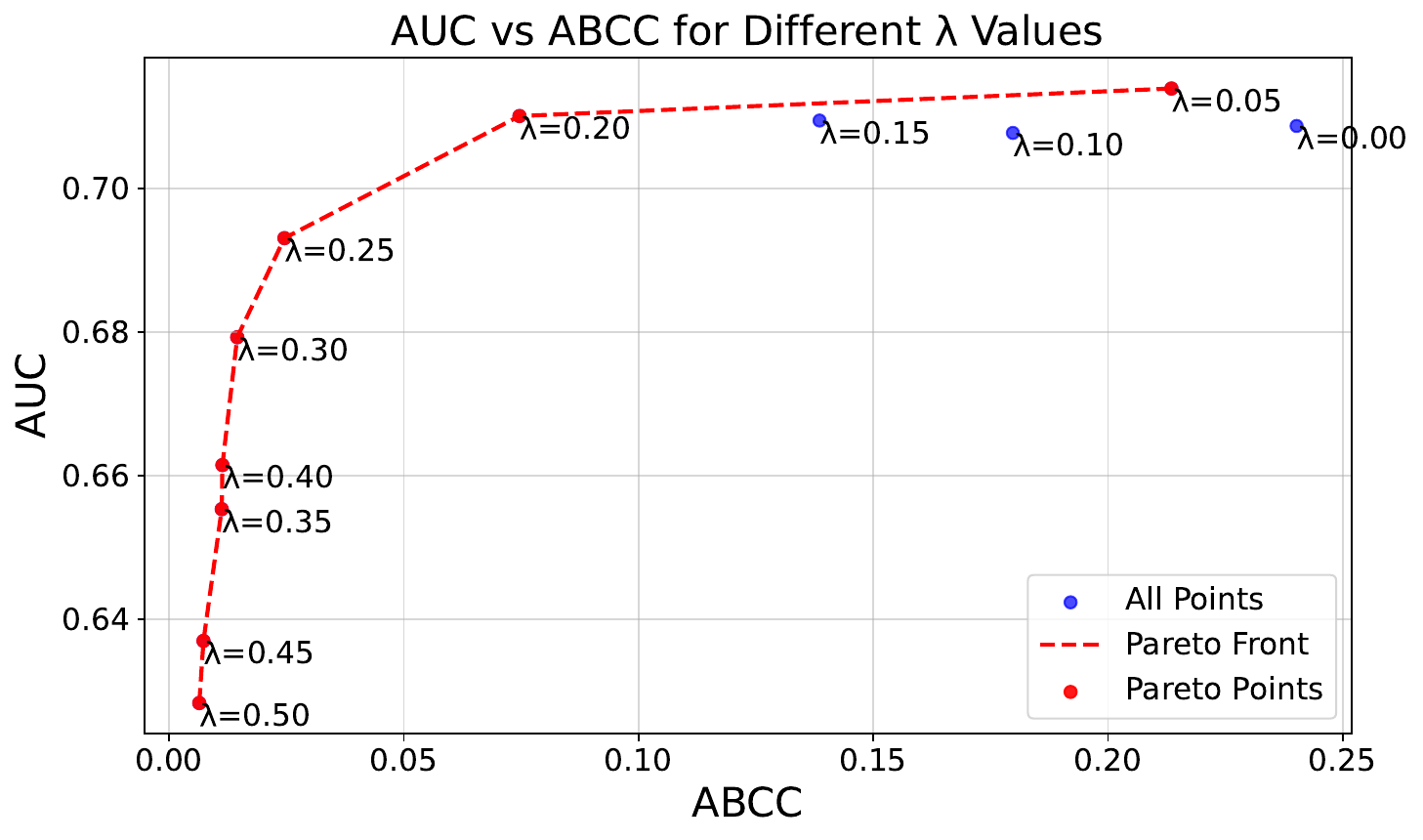}
\caption{AUC vs ABCC of lending\_high log.}
\end{subfigure}% \\

\begin{subfigure}[htbp]{0.49\linewidth}
\includegraphics[width=\textwidth]{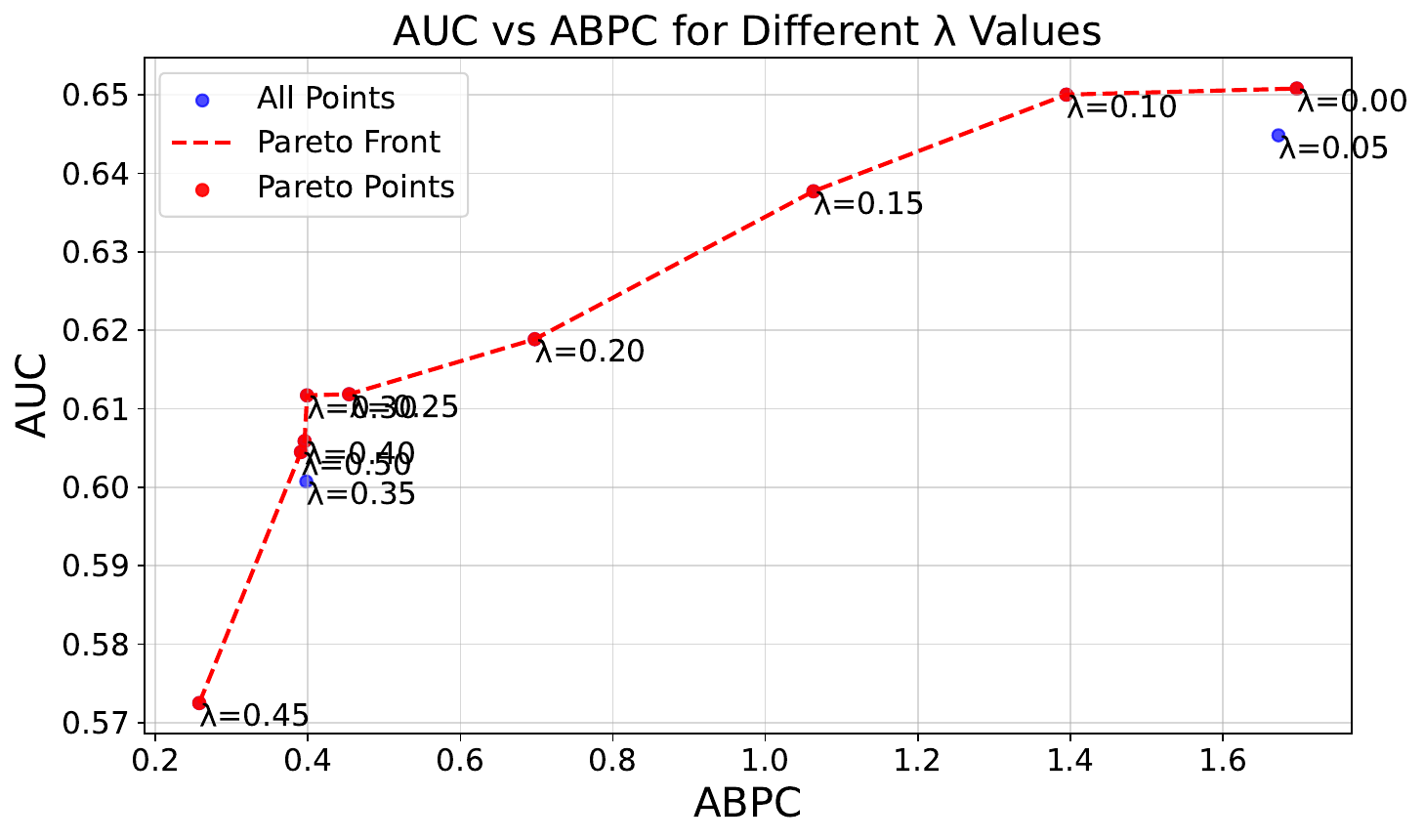}
\caption{AUC vs ABPC of renting\_high log.}
\end{subfigure}
\begin{subfigure}[htbp]{0.49\linewidth}
\includegraphics[width=\textwidth]{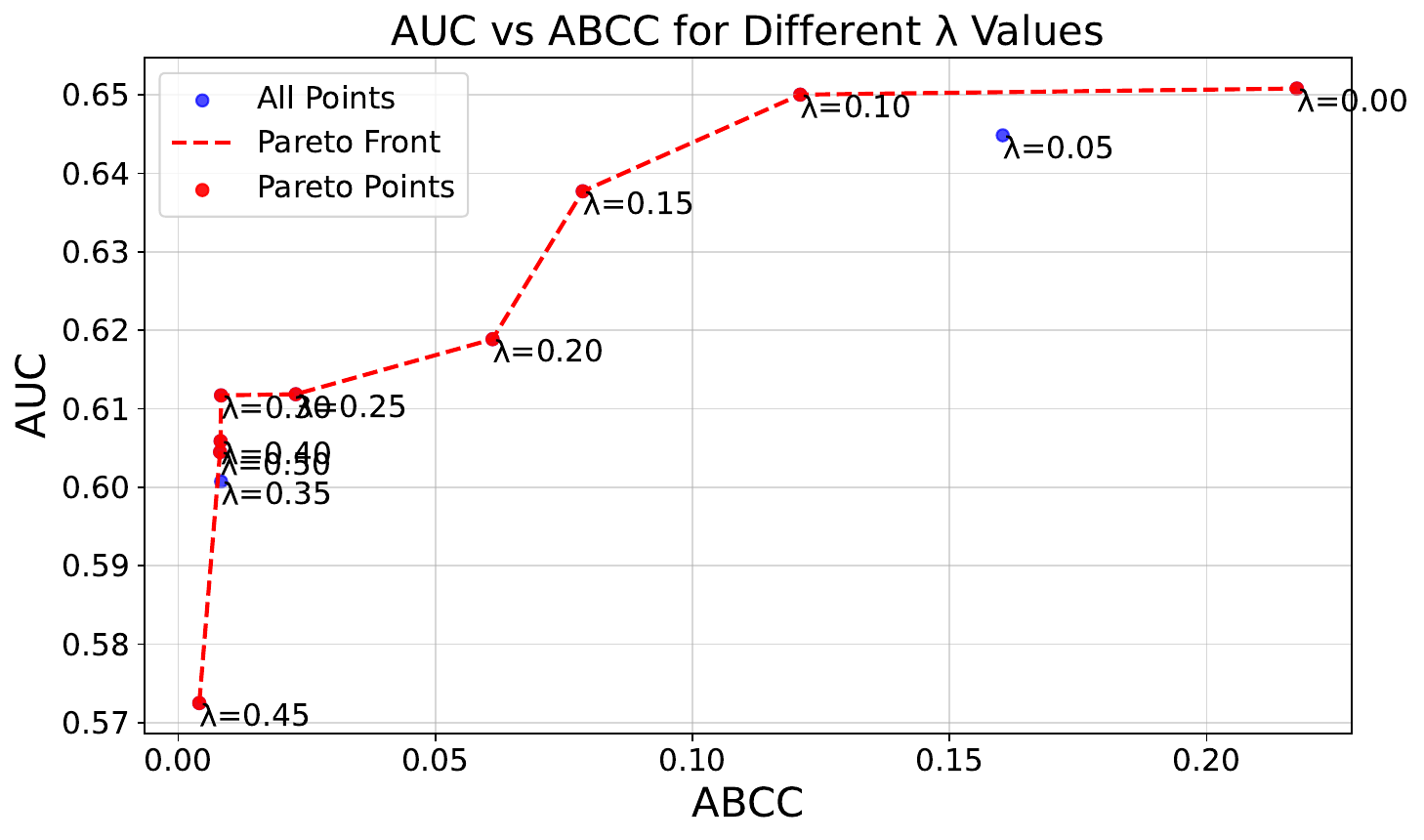}
\caption{AUC vs ABCC of renting\_high log.}
\end{subfigure}% 
\end{adjustwidth}
\caption{The results for including IPM loss with different values for $\lambda$.}
\label{fig:results_lambda}
\end{figure}

The second experiment evaluates the impact of incorporating the IPM loss, specifically the Wasserstein loss, into the training process. This experiment demonstrates how integrating this loss component reduces demographic parity violations while exploring its trade-offs with predictive performance. To achieve this, the setup uses the three event log variants with the highest bias levels. For each event log, classifiers are trained with varying $\lambda$ values, controlling the weight of the IPM loss relative to the BCE loss. The $\lambda$ values range from 0 to 0.5, incremented in steps of 0.05, providing a detailed exploration of its effects. The trained models undergo evaluation on the test set using two primary threshold-independent metrics: AUC to measure predictive capability and ABPC and ABCC to assess demographic parity violations. The results are presented in Figure~\ref{fig:results_lambda}. Next to a scatter plot, indicating the results for all values of $\lambda$, the Pareto points are indicated in red, and connected to show a Pareto front. For points not on the curve, there is at least one result for another $\lambda$ that scores better on both metrics. Since most points are either on the curve or relatively close to it, the results highlight the trade-off between predictive performance and fairness. Models with low $\lambda$ values achieve high AUC scores, reflecting strong predictive accuracy, but show significant demographic parity violations. In contrast, higher $\lambda$ values result in lower AUC scores, indicating reduced predictive performance, but successfully minimizing demographic parity violations, as measured by ABPC and ABCC.

\subsection{Discussion}

The results for both experiments 1 and 2 underscore the potential utility of the proposed metrics and the IPM loss function respectively. However, some limitations and challenges arise that warrant discussion. The experiments reveal that the predictive information contained in the prefixes may be limited, as evidenced by the low F1 scores and the shape of the propensity density curves. For instance, the propensity distributions in Figure~\ref{fig:distr_plots} lack the ideal bi-normal shape characteristic of binary classification. This suggests that class imbalance countermeasures or other probability adjustments may be necessary to optimize classifier performance. It also shows the limitations of using artificial data for classifier experiments. Nevertheless, since the primary objective of the experiments is to demonstrate proof-of-concept for the metrics and loss functions, these limitations do not undermine the core conclusions of the study.
The large observed difference between the $\Delta\mathit{DP}_b^{0.5}$ and $\Delta\mathit{DP}_b^{\text{opt.}}$ values, highlight the sensitivity of binary $\Delta\mathit{DP}$ to threshold selection. This underscores the advantage of adopting threshold-independent metrics like ABPC and ABCC. One remarkable observation from Tables~\ref{tab:results_exp1} and \ref{tab:results_exp1-removal} is the similarity between ABCC and $\Delta\mathit{DP}_c$ scores, which are often rounded to equal values. This similarity is in line with experimental results in the original work~\cite{han2023}. %The relatively non-overlapping nature of the propensity distributions in these experiments might be a contributing factor. 

\begin{figure}[htpb!]
\begin{adjustwidth}{-2.0cm}{-2.0cm}
\centering
\begin{subfigure}[htbp]{0.33\linewidth}
\includegraphics[width=\textwidth]{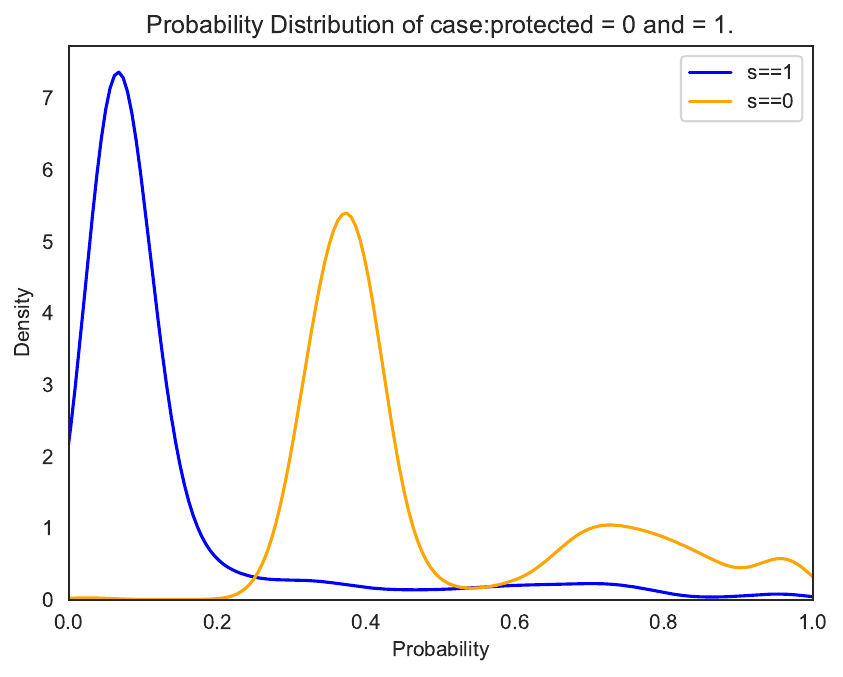}
\caption{$\lambda = 0.0$}
\end{subfigure}
\begin{subfigure}[htbp]{0.33\linewidth}
\includegraphics[width=\textwidth]{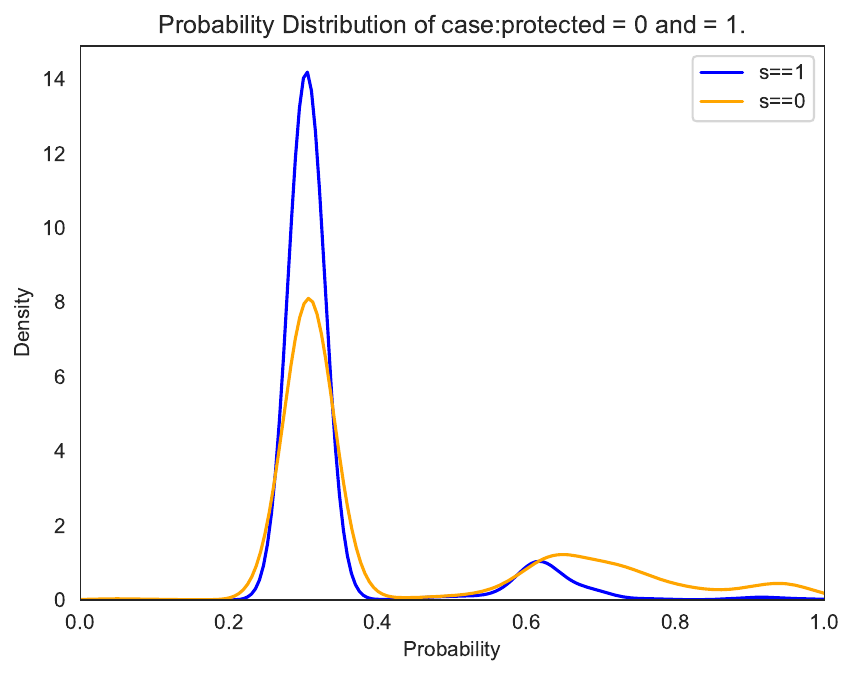}
\caption{$\lambda = 0.25$}
\end{subfigure}% 
\begin{subfigure}[htbp]{0.33\linewidth}
\includegraphics[width=\textwidth]{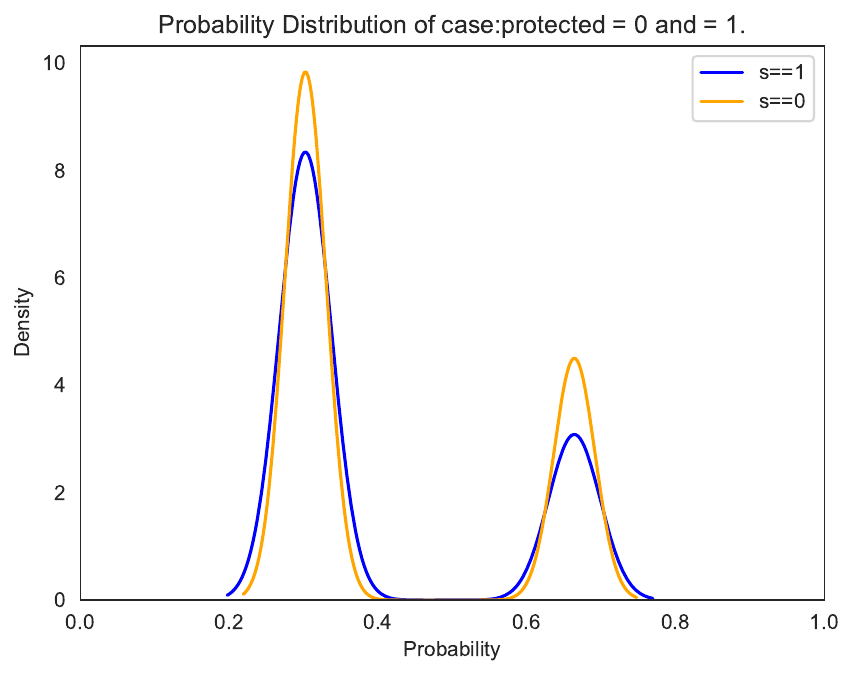}
\caption{$\lambda = 0.5$}
\end{subfigure}
\end{adjustwidth}
\caption{The LSTM propensity densities for the \textit{hiring\_high} event log.}
\label{fig:distr_plots_lambda}
\end{figure}

The results of Experiment 2, as illustrated in Figure~\ref{fig:results_lambda}, clearly demonstrate the trade-off between predictive quality and demographic parity violation when balancing the IPM and BCE loss functions. Most data points align along a clear Pareto front, with minor outliers remaining close to this frontier. Interestingly, for some event logs, small values of $\lambda$ yielded slight improvements in predictive performance compared to no IPM, potentially due to the regularization effect of IPM. To illustrate the effect of including the IPM loss on the PPM output, in Figure~\ref{fig:distr_plots_lambda} we plot the propensity distributions of the LSTM model for both protected groups, when trained on the \textit{hiring\_high} event log, with different values of $\lambda$ (0.0, 0.25, and 0.5). We can see the effect of increasing the importance of the IPM loss component, as the distributions of both groups show increased overlap. We also see that the shape of both distributions changes, which might additionally indicate a possible decrease in calibration (difference between propensity and true probability). 

\section{Practical Considerations}
\label{sec:practical}
Our work provides a practical framework for practitioners and researchers to assess and improve group fairness in OOPPM models using independence. Simpler methods such as threshold adjustments for different groups or batch-based groupings, are often insufficient for OOPPM. Processes often involve continuous data streams, and predictions are needed as early indicators for all running cases. We propose a principled approach to measuring and enhancing group-level independence using propensity scores or modifying the training process directly. Experimental results indicate that both $\Delta\mathit{DP}_c$ and distribution-based metrics ABPC and ABCC can effectively measure demographic parity violations (independent of the chosen threshold). Supported by qualitative reasoning in Section~\ref{sec:fairness} and the literature~\cite{han2023}, distribution-based metrics are recommended for a more robust evaluation. These metrics can be applied to any predictive model that outputs propensity scores, making them a low-effort addition to existing evaluation pipelines. Practitioners could use these metrics to check for potential biases, as they require only group identification labels and do not necessitate changes to testing workflows. It is important to note that the metrics should ideally be applied to the probability scores used for decision-making, even if these differ from the model's original propensity scores, such as in cases where calibration is applied. One key advantage of using threshold-independent metrics is flexibility. Threshold-independent metrics allow models to be evaluated for fairness while retaining the ability to adjust thresholds down the line for e.g. cost-sensitive purposes~\cite{vanderschueren2022predict}, without altering conclusions on demographic parity.

To achieve models with greater fairness across protected groups, incorporating the IPM loss into composite loss functions offers a practical trade-off between independence and predictive performance. In a production setting, an optimal choice for $\lambda$ could be determined based on validation result. Moreover, the trade-off can be tailored to the specific use case and its associated risks regarding independence fairness. However, practitioners may face challenges in choosing an optimal value without extensive validation. Future work could explore automated or adaptive tuning methods to alleviate this dependency. Our implementation of the Wasserstein distance as an IPM loss, built-in PyTorch\footnote{Version 2.5.1.}, is readily available to train compatible models (with a gradient-based optimization method). However, we recommend sufficiently large batch sizes (or full batch training) for distribution-based losses, as these rely on batch-level statistics. While computational constraints were not an issue in these experiments, scaling these methods to very large datasets (and batch sizes) could present challenges. Calculating Wasserstein distances or other IPMs for large datasets can be computationally intensive, though approximation techniques like Sinkhorn distances~\cite{Cuturi2013} can help.

It is important to note that this work assumes OOPPM models are deployed in contexts where process fairness has been, or is actively being improved. Without such improvements, deploying these models risks perpetuating existing biases rather than mitigating them. For instance, if OOPPM models are used to trigger early interventions that disproportionately keep cases from one protected (minority) group, the remaining cases for this group may exhibit an even lower fraction of positive outcomes compared to counterparts not belonging to that group and the inclusion of other fairness metrics is essential, such as separation or sufficiency. Furthermore, focusing exclusively on group independence during training might unintentionally reduce accuracy for certain groups, potentially even conflicting with other fairness definitions~\cite{barocas2023fairness}. Balancing these competing objectives requires careful consideration and alignment with the specific goals and risks of the application context.

As mentioned earlier, another important aspect to note is that model outputs (propensities) should only be interpreted as probabilities for well-calibrated models. Calibration depends on the classifier, and neural network models (e.g., LSTMs) have shown mixed results~\cite{Guo2017}. Miscalibration may worsen when models are trained using multi-objective criteria such as done here with the composite loss function, since fairness objectives can reshape output distributions at the cost of calibration. Look for example at the propensity distributions found in Figure~\ref{fig:distr_plots_lambda}, where increasing the value for $\lambda$ alters the shape of both distributions. Hence, decision-makers should interpret such outputs with caution, and further investigation on the effects of calibration on decision making in PPM would be interesting. 

\section{Conclusion and Future Work}
\label{sec:conclusion}

This work introduces group independence and demographic parity violation metrics to ensure fairness in OOPPM classifiers. In addition to traditional metrics like $\Delta\mathit{DP}$, which quantifies the average difference in classifier outputs between two protected groups, this study incorporates metrics based on probability density distributions. These novel metrics, inspired by recent advancements in machine learning~\cite{han2023}, offer a more robust threshold-independent evaluation of demographic parity. Furthermore, the study introduces the use of a composite loss function, including an IPM loss, specifically the Wasserstein distance, in combination with BCE loss. Both the applicability of the metrics and the effectivity of the adapted loss functions are shown in controlled experiments using LSTM neural networks. However, the methodology is designed to be compatible with a wide range of classifiers. By adjusting the weight of the IPM loss relative to the BCE loss, a Pareto front is formed, exploring the trade-off between improved predictive capabilities and reduced independence violations. This framework allows for informed decision-making, enabling stakeholders to select optimal trade-offs that align with varying application-specific fairness requirements. \\

Future research could expand this work in several directions. A first option involves incorporating other forms of group fairness, such as separation (error rate parity) and sufficiency (calibration fairness), or exploring individual fairness techniques, such as counterfactual methods. The proposed metrics and loss functions could also be adapted to handle continuous sensitive attributes or a combination of multiple (intersecting) sensitive attributes. Another possibility is to explore multiclass tasks, such as next-event or suffix prediction. Alternative strategies, including data augmentation, debiasing, or the use of generative models~\cite{deleoni_2024}, could also be evaluated using the fairness metrics introduced here. A more holistic approach might address related challenges such as class imbalance and label uncertainty for both outcomes and sensitive parameters~\cite{kim2021, Peeperkorn2023}. It is important to note that our experimental setup uses simplified, artificial datasets. While this choice aids in illustrating the core concepts of fairness measurement and loss function integration, real-world scenarios often involve complexities such as imbalanced data and multiple intersecting sensitive attributes. Addressing these challenges will be a key focus for future research.
%Expanding the scope of experiments to include real-world datasets and other types of classifiers could further validate and refine these methods, enhancing their practical utility in diverse predictive monitoring scenarios.

\begin{credits}
\subsubsection{\ackname}
This work was supported in part by the
Research Foundation Flanders (FWO) under Project 1294325N, and by the Flemish Government, through Flanders Innovation \& Entrepreneurship (VLAIO, project HBC.2021.0833).
%\subsubsection{\discintname}
\end{credits}
%\bibliographystyle{splncs04}
\bibliographystyle{splncs04nat}
\bibliography{bibliography}

\end{document}